\documentclass[letterpaper, 10 pt, journal, twoside]{IEEEtran}
\usepackage{subcaption}
\usepackage{amsmath,amsfonts}
\usepackage{algorithmic}
\usepackage{array}
\usepackage{textcomp}
\usepackage{stfloats}
\usepackage{url}
\usepackage{verbatim}
\usepackage{graphicx}
\usepackage{multicol}
\usepackage[bookmarks=true]{hyperref}
\usepackage{amssymb}  
\usepackage{multirow}
\usepackage{float}
\usepackage{booktabs}
\usepackage{xcolor}
\usepackage[normalem]{ulem}
\usepackage{cite}

\def\BibTeX{{\rm B\kern-.05em{\sc i\kern-.025em b}\kern-.08em
    T\kern-.1667em\lower.7ex\hbox{E}\kern-.125emX}}
\usepackage{balance}

\useunder{\uline}{\ul}{}

\begin{document}

\title{ConditionNET: Learning Preconditions and Effects for Execution Monitoring}

\author{Daniel Sliwowski$^{1}$ and Dongheui Lee$^{1,2}$
\thanks{$^{1}$Daniel Sliwowski and Dongheui Lee are with Autonomous Systems Lab, Technische Universität Wien (TU Wien), Vienna, Austria (e-mail: \texttt{\{daniel.sliwowski, dongheui.lee\}@tuwien.ac.at}).}%
\thanks{$^{2}$Dongheui Lee is also with the Institute of Robotics and Mechatronics (DLR), German Aerospace Center, Wessling, Germany.}%
}

\maketitle

\begin{abstract}
The introduction of robots into everyday scenarios necessitates algorithms capable of monitoring the execution of tasks. In this paper, we propose ConditionNET, an approach for learning the preconditions and effects of actions in a fully data-driven manner. We develop an efficient vision-language model and introduce additional optimization objectives during training to optimize for consistent feature representations. ConditionNET explicitly models the dependencies between actions, preconditions, and effects, leading to improved performance. We evaluate our model on two robotic datasets, one of which we collected for this paper, containing 406 successful and 138 failed teleoperated demonstrations of a Franka Emika Panda robot performing tasks like pouring and cleaning the counter. We show in our experiments that ConditionNET outperforms all baselines on both anomaly detection and phase prediction tasks. Furthermore, we implement an action monitoring system on a real robot to demonstrate the practical applicability of the learned preconditions and effects. Our results highlight the potential of ConditionNET for enhancing the reliability and adaptability of robots in real-world environments. The data is available on the project website: \href{https://dsliwowski1.github.io/ConditionNET_page}{https://dsliwowski1.github.io/ConditionNET\_page}.
\end{abstract}


\section{Introduction}
\IEEEPARstart{I}{n} recent years, there has been a desire to introduce robots into everyday scenarios like homes, elder-care facilities, or restaurants~\cite{ManipLearnRev}. These environments pose a new set of challenges as they are inherently unstructured, making it impossible to predict all the scenarios the robot will encounter at design time~\cite{ManipLearnRev}. This motivates the need for developing algorithms that can not only learn and adapt to new situations but also monitor the action execution, thus enhancing robustness. 

When manipulating our environments, we as humans, monitor the progression of our actions. For example, before pouring juice into a cup, we subconsciously check if we are holding a bottle and if the target cup is present. Similarly, post-pour, we confirm the juice is in the cup and that no spills happened. In robotics, the conditions that need to be satisfied before acting are named \textit{preconditions} and the ones holding after -- \textit{effects}. These conditions play an important role in monitoring executions as checking the preconditions facilitates re-planing~\cite{TPVQA, OON}, and checking for effects enables action re-execution or recovery~\cite{TPVQA}. However, the area of learning these conditions in a data-driven manner is fairly unexplored.

\begin{figure}[t]
    \centering
    \includegraphics[width=\linewidth]{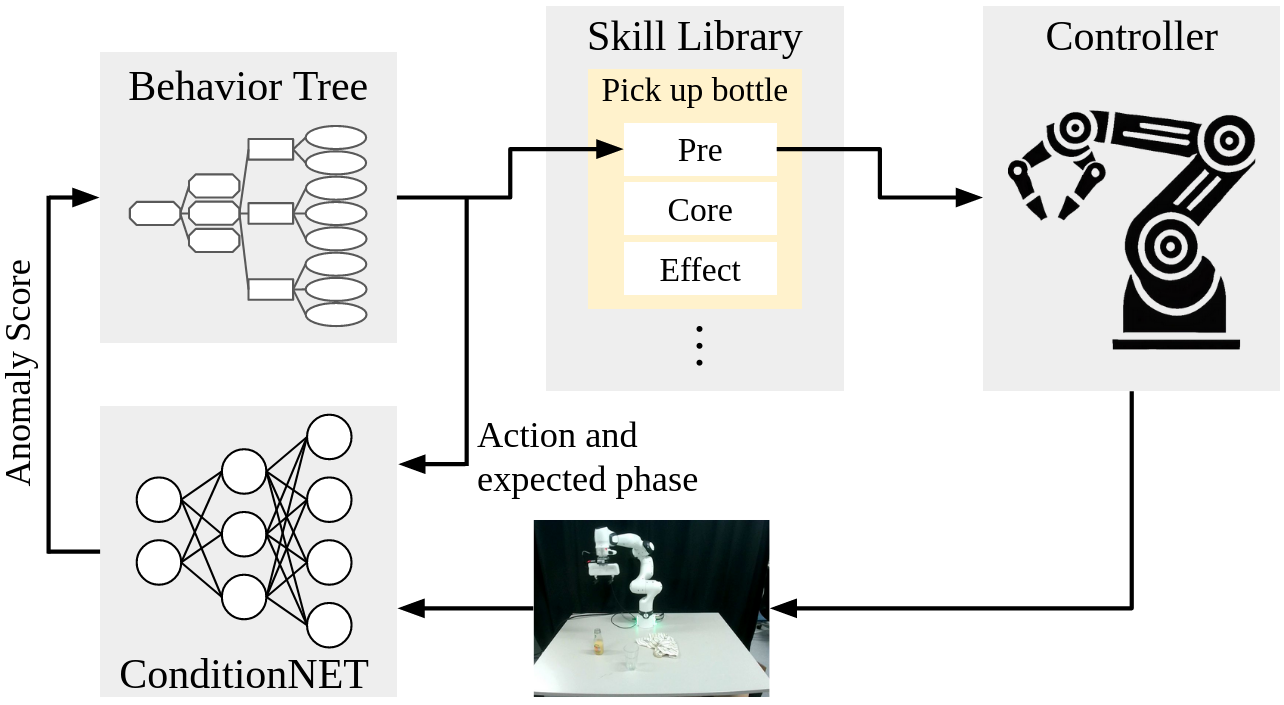}
    \caption{\textbf{An overview of the proposed anomaly detection and recovery algorithm.} ConditionNET detects anomalies by comparing the expected and current motion phases, the latter is predicted by a vision-language model. A behavior tree governs task execution, providing the action and expected phases, while motions are generated by a skill library and executed on the robot with an impedance controller.}
    \label{fig:overview}
    \vspace{-.4cm}
\end{figure}

Many studies focus on execution monitoring using internal sensor data such as force, torque, or end-effector pose~\cite{ANOMALY1, Eiband2019, CONT2}. However, their effectiveness is limited by their inability to determine the semantics of objects. For instance, consider a scenario where a robot is instructed to "pick up the bottle with apple juice," but mistakenly grabs a bottle of orange juice instead. If the robot relies solely on internal sensors, such as gripper state or force torque, it might incorrectly conclude that the action was performed correctly, since the gripper is closed and force is detected at the end effector. In reality, the action has failed because the wrong bottle was picked up. This limitation can be overcome by enabling the algorithm to understand the objects in the scene, which can be achieved by using exteroceptive sensors, such as cameras.

One open question is how to effectively convey the semantic information to the robot. Classical approaches use Planning Domain Definition Language (PDDL)~\cite{PDDL}, where each skill is described by a planning operator, which is parameterized by the objects involved in the action. However, specifying new skills can be challenging for non-experts due to the strict format required by planning operators. More recently, we have observed that natural language can serve as an input to task and motion planning algorithms, providing a more user-friendly interface. Moreover, works like CLIP~\cite{CLIP} and InstructBLIP~\cite{instructblip} demonstrate that, given sufficient data, an algorithm can generalize to previously unseen scenarios. 

To address these limitations, we propose learning the conditions in a vision-language manner. We formulate the anomaly detection and recovery problem by verifying whether the preconditions and effects of executed actions are satisfied. We learn these conditions in a data-driven manner using a transformer-based vision-language model, which we call ConditionNET. The goal is to predict, for a given action described in natural language, whether the current observation satisfies the precondition or effect of the action, or neither. Thanks to our fast inference time, we demonstrate that the model can be used for action execution monitoring by checking the conditions each time a new observation is available. Additionally, we exploit the connection between preconditions, effects, and actions. By observing changes between preconditions and effects, we can infer the action that took place. We propose explicitly incorporating this relationship into the learning process. When the learned representations align with this principle, we refer to them as 'consistent.' Figure \ref{fig:overview} illustrates our approach.

To summarize, the contributions of our paper are as follows:
\begin{enumerate}
    \item an efficient visual-language model for learning preconditions and effects of actions;
    \item a training approach for optimizing for consistency between the representation of preconditions, effects, and actions;
    \item a real-time execution monitoring system based on the learned action preconditions and effects;
    \item a real-world manipulation dataset of a robot performing bartending tasks that contain successful and failed executions.
\end{enumerate}

\section{Related Works}
\subsection{Execution Monitoring}
Over the years, as the interest in deploying robots in increasingly unstructured environments has grown, numerous execution monitoring and anomaly detection algorithms have been proposed~\cite{anomalydetroboticssurv}.

Among these approaches, one class focuses on probabilistic modeling of task executions. These methods rely on modeling successful task executions probabilistically, often using Gaussian Mixture Models (GMMs)~\cite{Eiband2019, willibald2020collaborative} or Gaussian Processes (GPs)~\cite{romeres2019anomaly}. They predict expected sensory measurements during task execution, which are then compared against current observations to identify anomalies. Alternatively, for detecting multimodal anomalies, Hidden Markov Models (HMMs) can be used, where the anomaly threshold is predefined~\cite{azzalini2020hmms}, or by estimating probabilistic thresholds based on the progression of the task execution~\cite{park2019multimodal}.

Probabilistic approaches are effective for modalities such as end-effector poses or force-torque measurements but face challenges with high-dimensional data such as images~\cite{ruan2011regularized}. An alternative approach is to use deep learning, which can extract meaningful features from high-dimensional data and simultaneously learn to detect anomalies~\cite{AnomalyDetSlip, FaluireClassification1, Altan2022CLUEAIAC, FinoNet, InnerMonologe, SuccessVQA}. Two commonly used methods include measuring the reconstruction error~\cite{AnomalyDetSlip} or directly classifying anomalies or successes~\cite{FinoNet, InnerMonologe}.

Anomaly detection approaches focus on predicting whether the action is going on according to plan or not, however, they do not reason about which actions can be performed from the current state. In contrast, ConditionNet determines the anomaly through the learned preconditions and effects of actions. In case of an anomaly, the preconditions can be used in the recovery system to limit the set of selectable actions to only those that are feasible, thus simplifying the recovery problem.

Once an anomaly is detected, several strategies can be adopted to recover from the error. Works like~\cite{ANOMALY1} employ a human-in-the-loop strategy, where the user is queried to resolve the anomaly by either refining the action or adding new recovery skills to the robot's knowledge. In cases where multiple solutions exist, Eiband et al.~\cite{Eiband2019} propose to select among all solutions except the one that led to the anomaly, the one that has the minimal Mahalanobis distance to the current state. Recent works like REFLECT~\cite{liu2023reflect} adopt Large Language models (LLMs) to plan the recovery behavior by providing the current state description and anomaly reason in the prompt.

In principle, ConditionNET can be applied to any of the aforementioned recovery strategies in place of the anomaly detector. In this work, we employ a Behavior Tree (BT)~\cite{BT} with a pre-planned task and recovery plan, where branches are conditionally executed or preempted based on anomaly prediction. 

\subsection{Learning Action Preconditions and Effects}
Symbolic planners have been a part of artificial intelligence for many decades, and languages like PDDL~\cite{PDDL} have been a staple of modeling actions. Each action is described through a ``planning operator'' which defines the preconditions and effects of the action. Various approaches try to learn these planning operators from symbolic states~\cite{CLASSICAL4, CLASSICAL5} by identifying recurring predicates in the state of the environment before and after an action. However, since they rely on symbolic variables, additional techniques~\cite{CONT2} are needed to estimate predicates from sensor readings.

Instead of learning to ground the predicates from the observations, recent learning approaches attempt to learn the preconditions and effects directly from the data. Approaches like SuccessVQA~\cite{SuccessVQA} and InnerMonologe~\cite{InnerMonologe}, focus solely on classifying whether an action has been performed successfully. They assume that the action is always possible to execute, which is often not the case in real-world scenarios. For example, if a robot is tasked with picking up a bottle and the bottle is removed during execution, these models can only detect the anomaly after the task is completed. In contrast, ConditionNET also reasons about the preconditions of actions. In the previous example, our model could detect that the bottle has been removed, and the precondition of the action is no longer satisfied. This allows the robot to react and preempt the motion execution to perform another action instead.

More recent trends in task planning utilize LLMs~\cite{PALME}. These models can effectively plan tasks using few or even no examples due to the massive amount of data they observed during training. By adding vision as an input modality, PALM-E~\cite{PALME}, a Large Visual-Language Model (LVLM), can reason about the environment's state from images. This allows checking preconditions by querying the model about the feasibility of an action based on the observation and verifying the action's success post-execution. Conceptually, this approach aligns with ours, as we also aim to determine if the current observation is the precondition or effect of an action. However, as these types of models are computationally intensive, they are not suited for real-time execution monitoring. Additionally, due to their size, LLMs and LVLMs are difficult to deploy on robots, especially those without internet access for cloud computing. In contrast, ConditionNET, with only 30M parameters is around 187 times smaller than PALM-E 562B and can be deployed relatively easily. An added benefit of the small size of the model is a considerable decrease in the inference time, which makes it feasible to run inference for each frame from the camera.

\begin{figure*}
    \centering
    \includegraphics[width=0.95\textwidth]{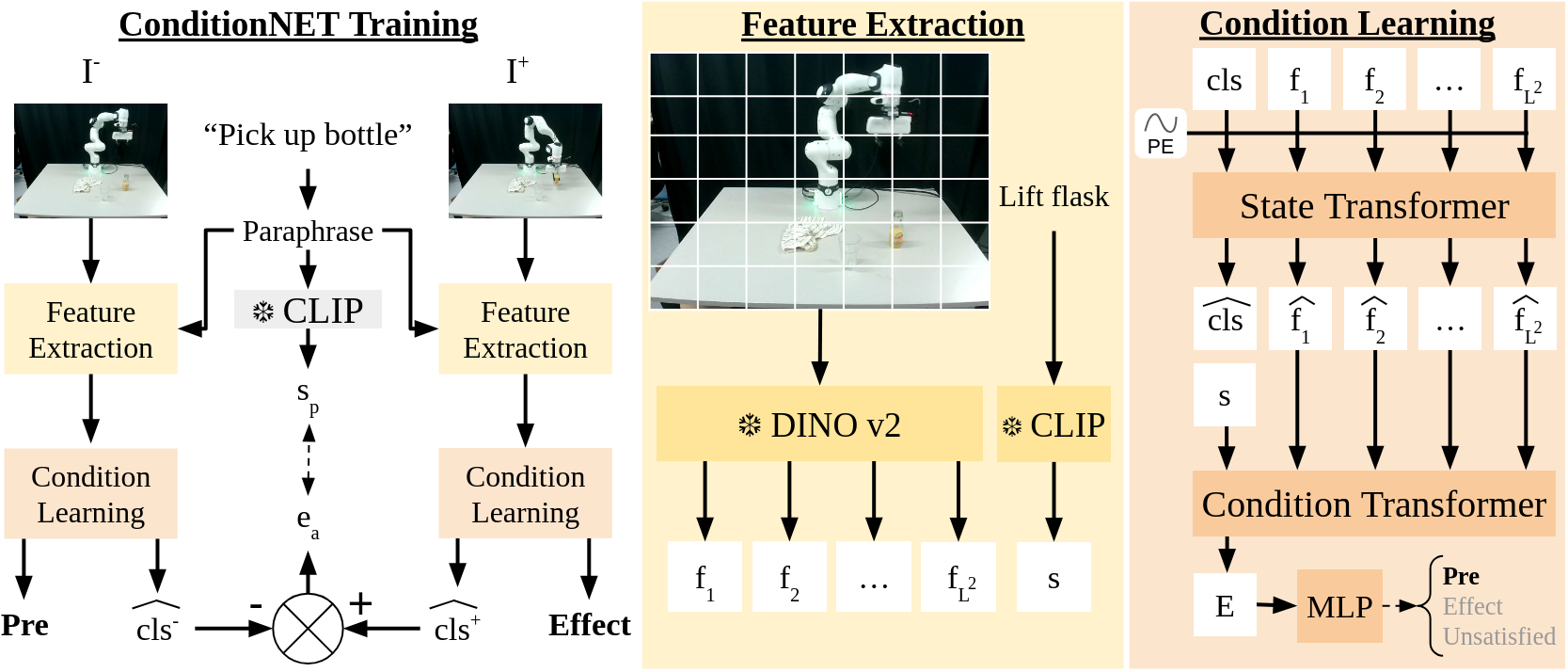}
    \caption{\textbf{ConditionNET Architecture.} For an image-action pair, we compute the condition feature \( E \) and classify the current observation as precondition, effect, or unsatisfied. We extract image and semantic features using DINOv2~\cite{Dinov2} and CLIP~\cite{CLIP}. The State Transformer then extracts the general state feature \( \widehat{cls} \), and the Condition Transformer extracts the condition feature \( E \). For consistency loss, we use features from both the precondition frame and the effect frame, denoted \( - \) and \( + \), respectively. We compute the action feature \( e_a \) as the difference between \( \widehat{cls^+} \) and \( \widehat{cls^-} \). Using InfoNCE loss~\cite{oord2018representation}, we make the action feature ``similar'' to the paraphrased action description \( s_p \), but only for successfully executed actions.
}
    \label{fig:architecture}
\end{figure*}

\section{Methodology}
\subsection{Problem formulation}
We formulate the task of learning the preconditions and effects of actions as a state prediction problem. Given a frame $I$ and a natural language description of action $a$, the goal is to classify whether $I$ is the precondition or effect of action $a$, or unsatisfied both of them.

We learn the action conditions from a dataset $\mathcal{D}$, which contains $N$ demonstrations of skill executions. Each demonstration $i$ consists of $M_i$ frames, a natural language description of the action $a$, a label indicating whether the action was successful, and a temporal segmentation mask, which splits the demonstration into three phases: the preparation phase, the core phase, and the post phase.

To optimize for consistent representations of the precondition and effect features, we use the pre-image $I^-$, sampled from the action preparation segment, and the effect-image $I^+$, sampled from the action post segment, along with the natural language description of the task $a$. We create the $(I^-, I^+, a)$ triplets by pairing all possible combinations of images from the preparation and post segments of a given demonstration.

\subsection{Data augmentation}
Due to data sparsity, we augment the training data by leveraging known relationships between actions in the dataset. Specifically, the post-state of one action can be used as the pre-state of another, and vice versa. For example, the pre-state of picking up a bottle (where the bottle is on the table and the gripper is empty) is equivalent to the post-state of placing the bottle on the table. Similar relationships exist among other actions in our dataset. We experimentally determined a value of 0.5, which yielded the best performance. 

Additionally, to improve the robustness of different ways of describing the action, we paraphrase the natural language description of the action. To this end, we query an LLM (ChatGPT-4o~\cite{chatgpt}) to generate 20 different variants for each action and object name. Next, from the generated variations, we remove ones that deviate too far from the original meaning. We append the original actions and object names to the variations list. Finally, for each action description, we randomly sample an alternative action formulation and object names.

\subsection{Model architecture}
The model architecture is shown in Fig. \ref{fig:architecture}. First, the model needs to understand the overall scene from the vision data. To achieve this, we use a pre-trained, frozen DINOv2~\cite{Dinov2} backbone that extracts local and global features from the image. DINOv2 divides the image into \(L \times L\) patches and computes a per-patch representation. Thus, the frame is summarized into a sequence of \(L^2\) tokens \(F = \left[f_1, f_2, \ldots, f_{L^2}\right] \in \mathbb{R}^{D \times L^2}\). The value of $L$ depends on the used version of DINOv2 and the image size; in our case, we use {\tt dinov2\_vits14} and an image size of 224 by 224 pixels, which results in $L = 16$. 

Although the backbone effectively captures the image content, it might not focus on the objects and their relations, which are crucial for learning the conditions. To address this, we split our model into two stages using separate multi-head self-attention transformers~\cite{vaswani2017attention}.

In the first stage, the State Transformer aims to obtain high-level features that encode the overall state of the environment, such as the position of the gripper, whether the gripper is closed or open, and the locations of objects. However, these general features might include information irrelevant to a specific action. For example, in Fig. \ref{fig:architecture}, the task is to "put the spatula on the cutting board," and the information that broccoli is on the table is not relevant. Therefore, we introduce a second stage, the Condition Transformer, which is conditioned on the action being performed. This allows the model to focus on information relevant to the action.

In both stages, we use a strategy similar to the Vision Transformer~\cite{ViT}. In the first stage, we prepend a learnable \(cls\) token to the features computed by the visual backbone: \(\left[cls, f_1, \ldots, f_{L^2}\right] \in \mathcal{R}^{D\times L^2+1}\), and use a sinusoidal positional embedding (PE) to inform the model of the patches' order. We denote the state features resulting from the first stage as \(\left[\widehat{cls}, \widehat{f_1}, \ldots, \widehat{f_{L^2}}\right]\). Due to the training procedure, \(\widehat{cls}\) can be understood as a global representation of the state of the image, with the remaining features as local state representations.

In the second stage, we replace the \(\widehat{cls}\) token with the action semantics \(s\), computed using a frozen CLIP~\cite{CLIP} text encoder. We use the output token of the second stage corresponding to this semantic token as the condition representation \(E\). Finally, we use a Multilayer Perceptron (MLP) to classify whether the image shows the pre-state, effect-state of the action, or does not satisfy either.

\subsection{Optimization objectives}
Our main training objective is to predict the action phase based on the current observation and action description. To achieve this, we minimize the Cross-Entropy loss between the output of the condition MLP $x$ and the ground truth phase $y$:

\begin{equation}
    \mathcal{L}_{\text{condition}} = - \frac{1}{B} \sum_{i=1}^{B} \sum_{j=1}^{3} \log\left(\frac{\exp(x_{i,j})}{\sum_{k=1}^3 \exp(x_{i,k})}\right) y_{i,j},
\end{equation}
where $B$ is the batch size.

To optimize for consistent action representations, we introduce additional optimization objectives. First, we compute the difference between the effect- and pre-state features, $\widehat{cls^+}$ and $\widehat{cls^-}$, and denote it as $e_a$. As it should be possible to determine what action was executed by observing the change in the state, we minimize the distance between $e_a$ and the semantic feature of the paraphrased action description $s_p$ using the InfoNCE loss~\cite{oord2018representation}:

\begin{equation}
    \begin{split}
    \mathcal{L}_{\text{consistency}} = & - \frac{1}{B} \sum_{i=1}^{B} \mathbf{1}(i) \log \frac{\exp(S_{ii} / \tau)}{\sum_{k=1}^{B} \mathbf{1}(k) \exp(S_{ik} / \tau)} \\
    & - \frac{1}{B} \sum_{j=1}^{B} \mathbf{1}(j) \log \frac{\exp(S_{jj} / \tau)}{\sum_{k=1}^{B} \mathbf{1}(k) \exp(S_{kj} / \tau)},
    \end{split}
\end{equation}
\begin{equation}
    S_{ij} = \frac{e_a^i \circ s_p^j}{\|e_a^i\| \|s_p^j\|},
\end{equation}
where $S_{ij}$ is the cosine similarity between feature $i$ and feature $j$, $\tau$ is a learned temperature parameter, and $\mathbf{1}(i)$ is an indicator function that determines whether the $i$-th data triplet is used in the loss computation. We use the indicator function to filter out demonstrations where the action was unsuccessful. Intuitively, the first term in the consistency loss maximizes the similarity between the action feature and its corresponding description while minimizing the distance to all other descriptions. The second term ensures the loss is symmetric, i.e., the same whether we do image-to-text or text-to-image.

The loss is a weighted sum of all the intermediate losses:
\begin{equation}
    \mathcal{L} = \alpha \mathcal{L}_{\text{condition}}  + \beta \mathcal{L}_{\text{consistency}},
\end{equation}
where $\alpha$ and $\beta$ are selected such that each loss is normalized to 1 after the first batch.

\begin{figure}[t]
    \centering
    \includegraphics[width=\linewidth]{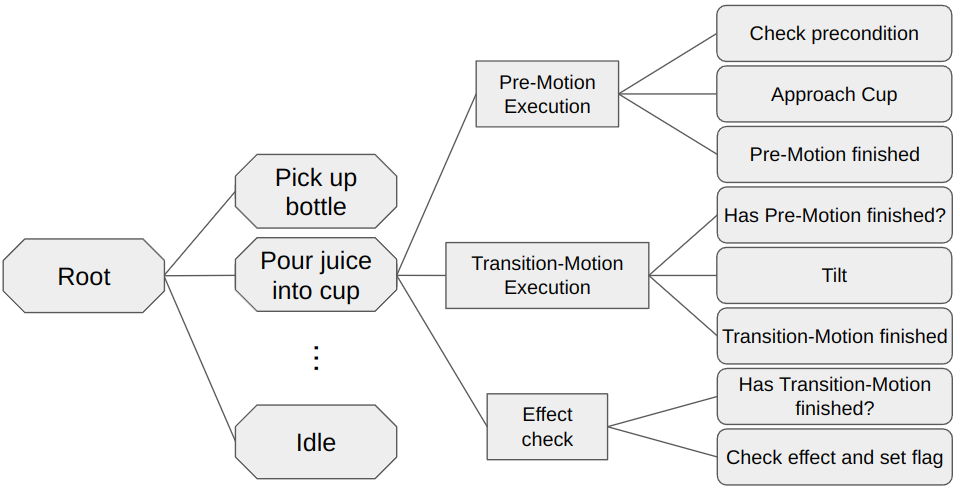}
    \caption{\textbf{Simplified depiction of the Behavior tree.} Octagons -- Selector nodes, rectangles -- sequence nodes, rounded rectangles -- behaviors.}
    \label{fig:bt}
    \vspace{-.2cm}
\end{figure}

\subsection{Training details}
We trained the model for 40 epochs, utilizing batches of 32 demonstrations per batch. The ADAMW optimizer was employed with a cosine annealing learning rate schedule, incorporating warm-up and reaching a peak learning rate of 0.0005. We set the optimizer's weight decay to 0.2, and used $\beta_1 = 0.9$ and $\beta_2 = 0.98$ for the momentum and squared gradient decay rates, respectively. Finally, the CLS token was randomly initialized by sampling from a Gaussian distribution of zero mean and unit variance.

\subsection{Motion Execution, Anomaly Detection and Recovery}
We build a library of skills that the robot can autonomously execute to complete tasks and handle failures. Each skill is composed of three phases: pre phase, core phase, and effect phase. The pre phase is associated with approach and preparation behaviors, the core phase involves completing the main part of the action, and the effect phase takes place after the tasks are executed when the effects should be observable. We associate each pre and core phase with motion primitives, which generate the motions to complete the phase; in the effect phase, we only check for successful action completion. In our case, we plan the motions by linearly interpolating between the start and end poses, though any other motion planning approach can be used.

To execute the given task, we use a Behavior Tree (BT) to establish the skill execution order. The BT selects the appropriate action and its phase from the library, given the current observation. Figure~\ref{fig:bt} illustrates a simplified version of the BT used in our experiments.

ConditionNET learns the action preconditions and effects rather than anomalies directly, necessitating additional steps for anomaly detection. Anomalies are detected by comparing the expected states of motion, given by the behavior tree, to those predicted by the model. During the pre phase, the expected state is the precondition, and any deviation is considered an anomaly. Similarly, in the effect phase, the expected state is the effect, and deviations indicate anomalies. During the core phase, where the current state may be ambiguous (e.g., during pouring or wiping), anomaly detection is suspended.

If an anomaly is detected during the pre phase, the current action halts, and the next action in priority is executed. If no actions can be executed, we fall back to the idle behavior. During the effect phase, the system verifies if the action has been successfully executed. This information is used to trigger recovery behavior branches when needed.

Although the model performs well overall, we still encounter some false positive anomaly detections. To address this issue, we added a filtering step to the anomaly scores. Specifically, for an anomaly to be considered valid, it must be detected across 8 consecutive frames.

\section{Experiments}
\subsection{Dataset and Metrics}
\textbf{FAILURE}~\cite{FinoNet} contains 235 demonstrations of Baxter robot executing skills such as placing, pouring, moving, pushing, and putting objects in or on other objects. The dataset includes both successful and unsuccessful executions. We augment the annotations with natural language descriptions of the actions and temporal segments into pre- and effect-states. The action descriptions follow these templates: {\tt place \(O_1\) on \(O_2\)}, {\tt pour balls into \(O_1\)}, {\tt push \(O_1\)}, {\tt put \(O_1\) on \(O_2\)}, {\tt put \(O_1\) in \(O_2\)}, where \(O_1\) and \(O_2\) represent the names of the objects manipulated.

\textbf{(Im)PerfectPour} is a teleoperated dataset collected in the Autonomous Systems Lab at TU Wien. It includes 406 successful and 138 unsuccessful demonstrations of a Franka Emika Panda robot performing tasks such as picking, placing, pouring, and whipping. The dataset consists of 4 skills: {\tt pick up \(O_1\)}, {\tt pour \(O_1\) into \(O_2\)}, {\tt place \(O_1\) on \(O_2\)}, {\tt wipe \(O_1\)}, where \(O_1\) and \(O_2\) are chosen from \texttt{bottle}, \texttt{juice}, \texttt{cup}, \texttt{table}, \texttt{cloth}. Each demonstration is recorded using two cameras to increase dataset variability. 
We assume the robot is the only agent affecting the scene, with anomalies being visually distinguishable. After recording demonstrations, we annotate each action's name, pre-, core-, and post-segments, along with any anomalies (e.g., spills or missing objects). We consider an execution anomalous if an action fails or cannot proceed. During labeling, we assume each action’s effect segment overlaps with the next action’s pre-segment.

\subsection{Baselines}
\textbf{FinoNET}~\cite{FinoNet} -- FinoNET focuses solely on anomaly detection. It classifies a window of 8 frames uniformly sampled from the action execution as either successful or failed. In our experiments, we use the official implementation provided by the authors.

\textbf{TPVQA}~\cite{TPVQA} -- TPVQA tackles anomaly detection as a visual question-answering problem (VQA). For each predicate present in the PDDL description of the skill planning operator, a corresponding question is constructed. These questions are used to query the ViLBERT~\cite{Lu2019ViLBERTPT} model. We use the same implementation of ViLBERT provided by AllenNLP~\cite{AllenNLP} without fine-tuning on our dataset. Questions about predicates are manually prepared in a manner consistent with the one presented in the paper.

\textbf{CLIP + MLP} -- We propose one additional baseline that simply predicts the phase based on concatenated CLIP image and text features with a Multilayer Perceptron (MLP). We only train the MLP, and the CLIP encoders~\cite{CLIP} are frozen.

\subsection{Quantitative Results}
We evaluate all models on both the FAILURE~\cite{FinoNet} and (Im)PerfectPour datasets. During training, we apply the paraphrasing and text augmentations for the baselines. We apply a 70/30 split of the data into the training and validation set, maintaining the distribution of failed and successful executions. During the evaluation, the action descriptions are kept fixed, i.e., no paraphrasing or augmentations are used. We report the accuracy, precision, recall, and F1 scores for both the anomaly detection task and inferring the phase of the action. As FinoNET~\cite{FinoNet} only focuses on the task of anomaly detection, we do not report the metrics for the latter task. As seen in Tab.~\ref{tab:quantitative}, ConditionNET outperforms the state-of-the-art baselines by a margin of at least 10\% for the anomaly detection task and 44\% for the action phase prediction task.

TPVQA has the worst overall performance on both tasks and datasets. During the evaluation, we noticed that general visual-question-answering models, like VILBERT~\cite{Lu2019ViLBERTPT}, are capable of correctly answering questions about the object relationships but fail to do so with questions about the robot itself. For example, whether the robot is holding an object or not. As these types of models are trained on datasets of typical, daily scenarios, images with robots in them are rare. This, in turn, results in the models not being able to properly understand the embodiment of the robots resulting in an inability to answer questions about them.

FinoNET does not condition the model predictions on the action that is being performed. This results in a more difficult learning task, as the model, additionally to inferring whether the action was successful or not, has to learn what action is being performed. On the other hand, in ConditionNET, we assume access to the action that the robot is performing and use it to condition our prediction. This allows the model to focus on aspects of the input data that are most relevant to the performed action. Note that, this is not a strong assumption, as the robot usually knows what action it is performing.

The CLIP+MLP and DINO+MLP baselines only use an MLP for learning the action conditions. This is a simple network architecture and it lacks sufficient expressiveness to properly learn the action conditions. On the other hand, in ConditionNET, we use two additional transformers to enhance the raw image features and better encode the action conditions.

We compare the size and inference time of our approach with standard VQA and LVLMs. We use a PC with an RTX A4000 GPU and an Intel 12700K CPU. For parameter count, we include only network components that process the data each timestep (i.e. without the CLIP model, as language features are extracted once). ConditionNET has about 30M parameters, much smaller than Flamingo3B (3 billion) and PALM (562 billion). For inference time over 1,000 batches (each batch evaluating 6 actions), ConditionNET achieves 21 $\pm$ 18 ms, significantly faster than Flamingo3B's~\cite{flamingo} 1493 $\pm$ 124 ms and ChatGPT-4's 14100 $\pm$ 2900 ms, making ConditionNET at least 71 times faster.

\begin{table}[t]
\footnotesize
\centering
\caption{Quantitative evaluation.}
    \label{tab:quantitative}
\begin{tabular}{l@{\hspace{8pt}}c@{\hspace{8pt}}c@{\hspace{8pt}}c@{\hspace{8pt}}c@{\hspace{8pt}}c@{\hspace{8pt}}c@{\hspace{8pt}}c@{\hspace{8pt}}c@{\hspace{8pt}}}
\toprule
\multicolumn{9}{c}{\textbf{FAILURE}~\cite{FinoNet}} \\
\midrule
                            & \multicolumn{4}{c}{Anomaly Detection}                         & \multicolumn{4}{c}{Condition Learning}                        \\
\cmidrule(lr{0.5em}){2-5} \cmidrule(lr{0.5em}){6-9}
Model                       & Acc      & Pre     & Rec        & F1            & Acc      & Pre     & Rec        & F1            \\
\midrule
CLIP+MLP                    & 0.81    & 0.81    & 0.81          & {\ul 0.81}    & {\ul 0.8}     & {\ul 0.77}    & {\ul 0.71}    & {\ul 0.74}    \\
DINO+MLP                    & {\ul 0.82}          & {\ul 0.82}          & {\ul 0.84}          & 0.8           & 0.78          & 0.72          & 0.67          & 0.69          \\
FinoNET~\cite{FinoNet}      & 0.79          & 0.79          & 0.79          & 0.79          & -             & -             & -             & -             \\
TP-VQA~\cite{TPVQA}         & 0.62          & 0.67          & {\ul 0.82}    & 0.73          & 0.44          & 0.75          & 0.24          & 0.37          \\
ConditionNET                & \textbf{0.89} & \textbf{0.91} & \textbf{0.89} & \textbf{0.88} & \textbf{0.88} & \textbf{0.85} & \textbf{0.79} & \textbf{0.82} \\
\midrule
\multicolumn{9}{c}{\textbf{(Im)PerfectPour}}  \\
\midrule
CLIP+MLP                    & {\ul 0.86}    & {\ul 0.91}    & 0.86          & {\ul 0.87}    & {\ul 0.93}    & {\ul 0.79}    & {\ul 0.77}    & {\ul 0.78}    \\
DINO+MLP                    & 0.72          & 0.88          & 0.72          & 0.74          & 0.85          & 0.74          & 0.7           & 0.72          \\
FinoNET~\cite{FinoNet}      & 0.74          & 0.80          & 0.74          & 0.74          &  -            & -             & -             & -             \\
TP-VQA~\cite{TPVQA}         & 0.76          & 0.81          & {\ul 0.9}     & 0.85          & 0.44          &  0.74         & 0.17          & 0.27          \\
ConditionNET                & \textbf{0.97} & \textbf{0.97} & \textbf{0.97} & \textbf{0.97} & \textbf{0.99} & \textbf{0.98} & \textbf{0.97} & \textbf{0.97} \\
\bottomrule
\end{tabular}
\end{table}

\subsection{Ablation Studies}
We performed ablation studies to verify each component's contribution to our architecture. Given that both datasets are relatively small and prone to overfitting, we created an extended dataset by combining demonstrations from both the FAILURE~\cite{FinoNet} and (Im)PerfectPour datasets to better showcase the benefits of our approach. We report the same evaluation metrics as in the quantitative evaluation of our model, and they can be found in Tab.~\ref{tab:ablation}.

\begin{table}[t]
\footnotesize
\centering
\caption{Ablation studies (FAILURE~\cite{FinoNet} + (Im)PerfectPour). Trained and validated with paraphrasing.}
    \label{tab:ablation}
\begin{tabular}{l@{\hspace{8pt}}c@{\hspace{8pt}}c@{\hspace{8pt}}c@{\hspace{8pt}}c@{\hspace{8pt}}c@{\hspace{8pt}}c@{\hspace{8pt}}c@{\hspace{8pt}}c@{\hspace{8pt}}}
\toprule
\multirow{2}{*}{Variant} & \multicolumn{4}{c}{Anomaly Detection} & \multicolumn{4}{c}{Condition Learning} \\
 \cmidrule(lr{0.5em}){2-5} \cmidrule(lr{0.5em}){6-9}
                                   & Acc    & Pre    & Rec    & F1   & Acc    & Pre    & Rec    & F1    \\
\midrule
w/o State       & & & & & & & & \\ 
Transformer     & {\ul 0.92}                    & 0.92                    & {\ul 0.92}                    & 0.92                   & 0.95                    & {\ul 0.92}                    & {\ul 0.91}                    & 0.91                   \\
w/o Condition & & & & & & & & \\
Transformer & 0.62                    & 0.73                    & 0.62                    & 0.63                   & 0.74                    & 0.67                    & 0.68                    & 0.68                   \\
w/o $L_{consistency}$     & {\ul 0.92}                    & {\ul 0.93}                    & {\ul 0.92}                    & {\ul 0.93}                   & {\ul 0.96}                    & {\ul 0.92}                    & {\ul 0.91}                    & {\ul 0.92}                   \\
ConditionNET & \textbf{0.96}                    & \textbf{0.97}                    & \textbf{0.96}                    & \textbf{0.96}                   & \textbf{0.97}                    & \textbf{0.92}                    & \textbf{0.92}                    & \textbf{0.92}      \\            
\bottomrule
\end{tabular}
\end{table}

\begin{table}[t]
\footnotesize
\centering
\caption{Robustness studies (FAILURE~\cite{FinoNet} + (Im)PerfectPour). Validated with paraphrased action texts.}
    \label{tab:rob}
\begin{tabular}{l@{\hspace{8pt}}c@{\hspace{8pt}}c@{\hspace{8pt}}c@{\hspace{8pt}}c@{\hspace{8pt}}c@{\hspace{8pt}}c@{\hspace{8pt}}c@{\hspace{8pt}}c@{\hspace{8pt}}}
\toprule
Paraphrasing & \multicolumn{4}{c}{Anomaly Detection} & \multicolumn{4}{c}{Condition Learning} \\
\cmidrule(lr{0.5em}){2-5} \cmidrule(lr{0.5em}){6-9}
during training?                         & Acc    & Pre    & Rec    & F1   & Acc    & Pre    & Rec    & F1    \\
\midrule

No   & 0.92                    & 0.92                    & 0.92                    & 0.92                   & 0.95                    & 0.90                    & 0.88                    & 0.89                   \\
Yes   & \textbf{0.96}                    & \textbf{0.97}                    & \textbf{0.96}                    & \textbf{0.96}                   & \textbf{0.97}                    & \textbf{0.92}                    & \textbf{0.92}                    & \textbf{0.92}      \\
\bottomrule
\end{tabular}
\end{table}

We considered three model variants. First, the ``w/o State Transformer'' variant, consists only of the Condition Transformer with twice the depth to maintain a similar number of trainable parameters and does not use the consistency loss. Second, the ``w/o Condition Transformer'' variant only uses the State Transformer, and also does not use the consistency loss. Third, the ``w/o $L_{consistency}$'' variant, removes the $L_{consistency}$ loss from the training process. Additionally, we evaluated the influence of paraphrasing on model robustness by training one variant using a fixed set of action descriptions, whose results we present in Tab.~\ref{tab:rob}.

When using the consistency loss, we observed better model performance. The inclusion of this loss encourages the state transformer to extract better state features, focusing on aspects of the scene relevant to the actions. With higher-quality state features, the condition transformer learns improved condition features, leading to better phase prediction and anomaly detection performance.

The model performance when introducing the action in the middle of the transformer architecture (w/o $L_{consistency}$) compared to when introducing it at the beginning (w/o State Transformer) is comparable. The ``w/o Condition Transformer'' variant performs significantly worse than the proposed ConditionNET. Without the Condition Transformer, the model lacks information about the robot's action and cannot focus on key states in the image, highlighting the importance of conditioning the model on the robot’s current action.

To further test robustness, we trained the model with paraphrased action descriptions. The performance of the model on the original validation dataset provides an upper bound on its capabilities, as this set and task are the simplest. Without varied action descriptions, the model's performance declines when the actions are described differently. However, incorporating diverse formulations during training significantly improves performance on the modified validation set, nearly reaching the upper bound. This demonstrates that training with varied action descriptions enhances the model's robustness.

\begin{figure*}[t]
    \centering
    \includegraphics[width=\textwidth]{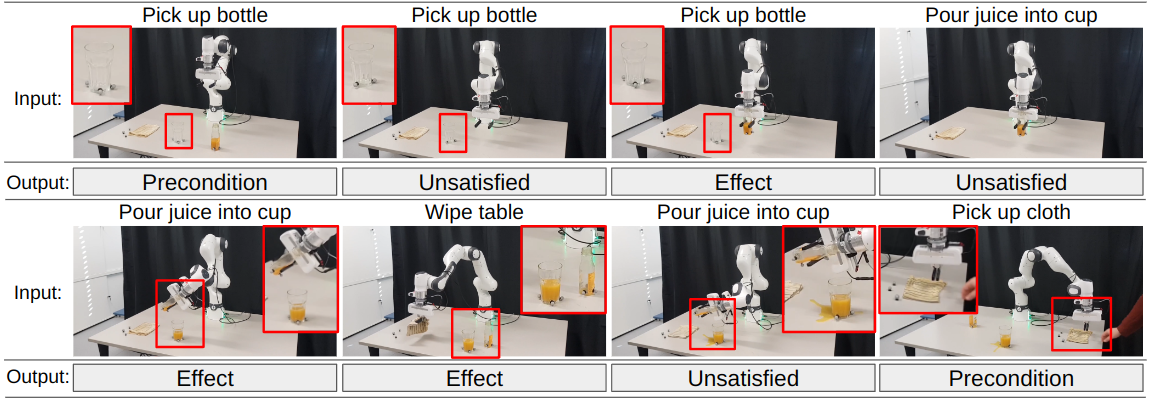}
    \caption{\textbf{Qualitative Results} show the model performance in the continuous action monitoring experiment. For clarity of presentation, only the results for single actions have been shown, but in reality, predictions for all actions are made in parallel. We highlight less visible objects in the red boxes.}
    \label{fig:monitor}
\end{figure*}

\begin{figure*}
    \centering
    \includegraphics[width=.9\linewidth]{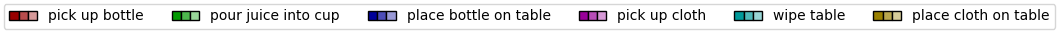}
    \begin{subfigure}{.5\textwidth}
      \centering
      \includegraphics[width=.95\linewidth]{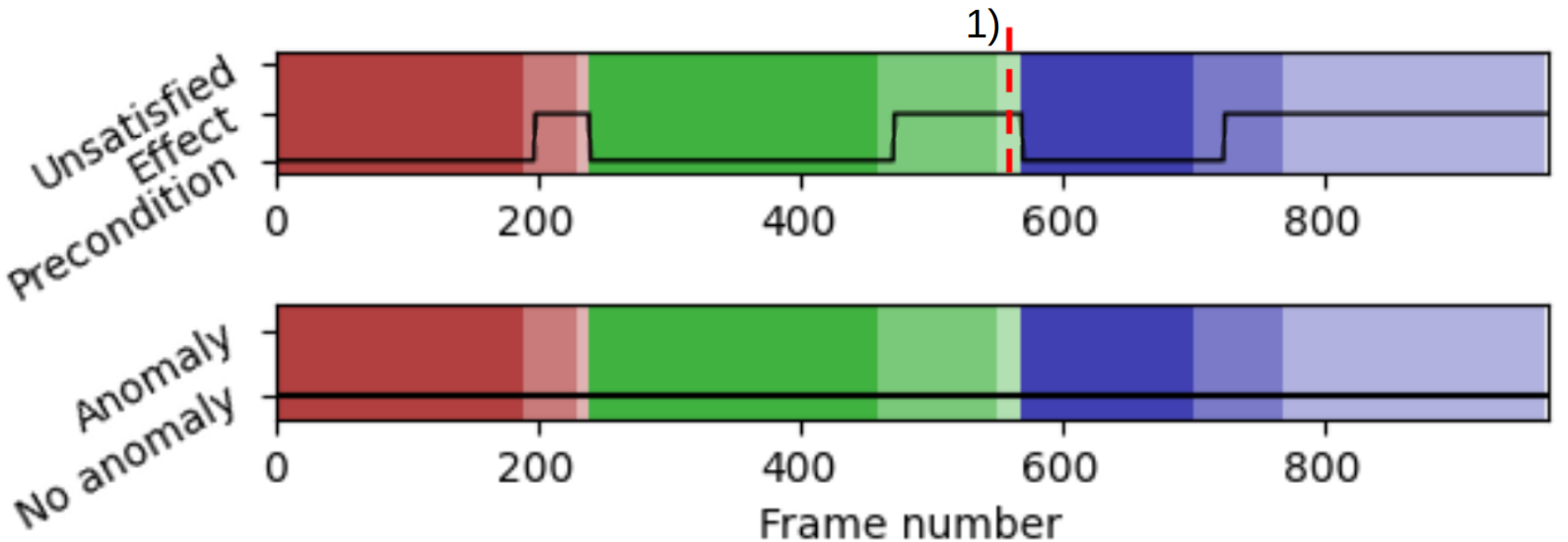}
      \caption{No anomalies during execution}
      \label{fig:nospill}
    \end{subfigure}%
    \begin{subfigure}{.5\textwidth}
      \centering
      \includegraphics[width=.95\linewidth]{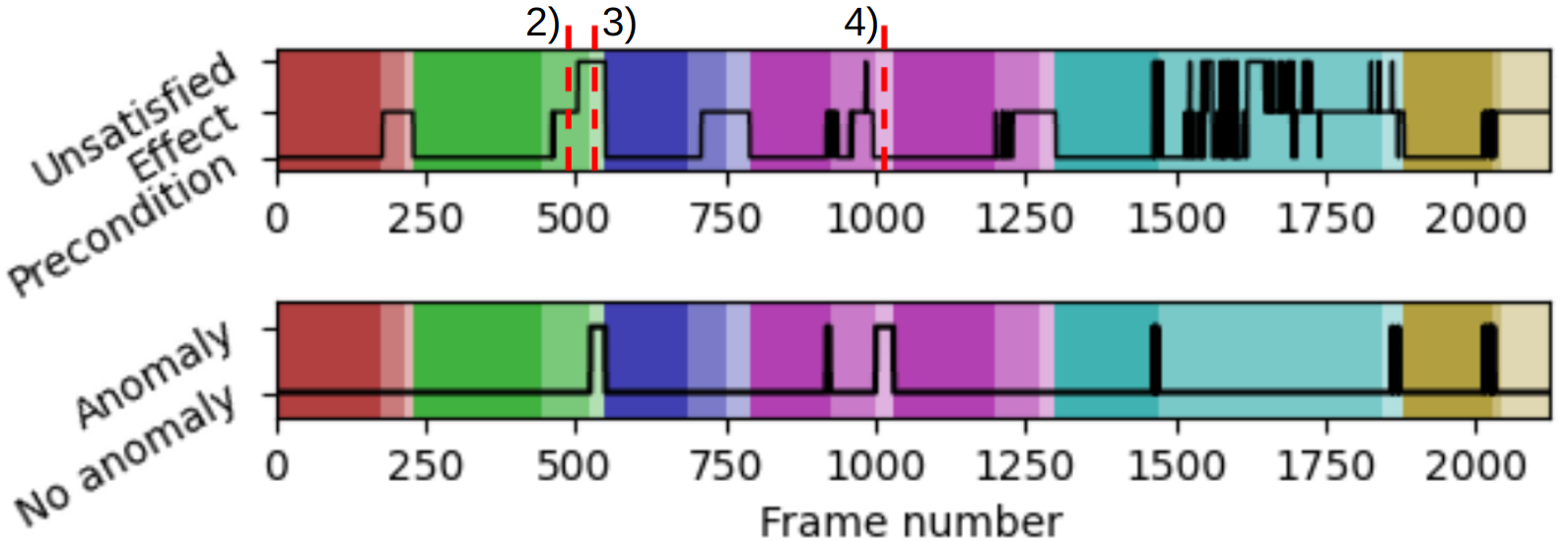}
      \caption{Anomalies present during execution}
      \label{fig:spill}
    \end{subfigure}
    \begin{subfigure}{\textwidth}
      \centering
      \includegraphics[width=\linewidth]{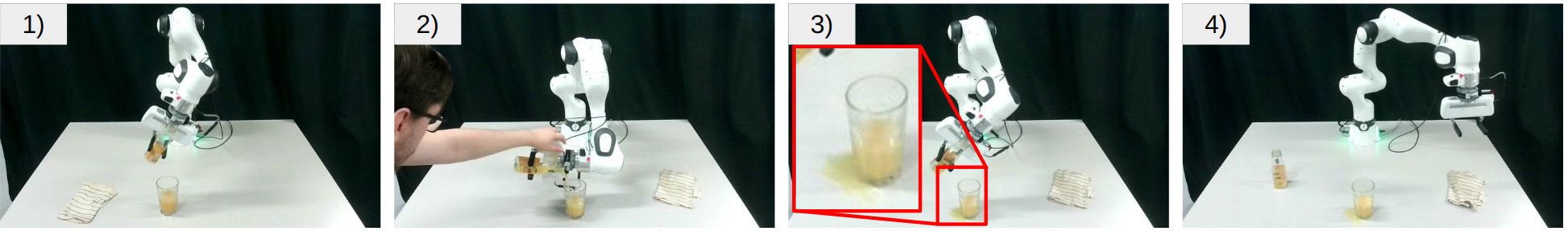}
      \caption{Snapshots form execution}
      \label{fig:snapshot}
    \end{subfigure}
    \caption{Phase prediction confidences and anomaly prediction results over time. Each hue marks a different expected action, and each saturation denotes a different expected motion phase. The most saturated color represents the pre-motion, the medium saturation denotes the core-motion, and the least saturated indicates the post-motion. The images below show snapshots for different points in the execution.}
    \label{fig:timeplots}
\end{figure*}

\subsection{Qualitative Results}
In this subsection, we demonstrate the performance of the proposed execution monitoring system. Fig.~\ref{fig:monitor} shows snapshots from the experiment. The first six images illustrate that the model correctly reasons about the presence or absence of objects and the successful execution of skills. In the seventh image, a human disrupts the robot during pouring, causing a spill. The algorithm accurately detects the spill and correctly identifies that the effects of the action are not met. The final image shows a frame taken just after a human has pushed the robot, preventing it from picking up the cloth. Here, the model correctly determines that the robot is still in the preconditioned state of the action and attempts to grab the cloth again.

Additionally, Fig.\ref{fig:timeplots} illustrates phase prediction and anomaly detection results over time. In Fig.\ref{fig:nospill}, the framework accurately predicts motion stages and identifies the absence of anomalies during task execution. In Fig.~\ref{fig:spill}, two anomalies occur. Around frame 500, a human pulls the robot, causing a spill. At the time of the incident, the model switches from predicting "effect" to "unsatisfied." Since the expected state is undefined during core motion execution, no anomaly is immediately reported. Once the expected phase changes to "effect," the anomaly is detected, triggering execution of the recovery branch.

The second anomaly occurs when the robot fails to pick up the cloth correctly in its initial attempt. The model correctly determines that the execution is still in the precondition state, but the expected state is "effect", which is detected as an anomaly. Subsequently, the robot retries and successfully picks up the cloth.

For a more comprehensive demonstration of the algorithm's performance, we refer the reader to the supplementary video.

\section{Conclusions}
In this work, we introduce ConditionNET, an algorithm for data-driven learning of action preconditions and effects for execution monitoring. We frame the learning task as a state prediction problem, where we predict whether the current observation-action pair satisfies the precondition, represents the effect of the action, or fails to meet either condition. By explicitly modeling dependencies between pre-state, effect-state, and action, we demonstrate improved model performance. ConditionNET consistently outperforms all baselines across various datasets by a margin of at least 6\%, despite having only 30 million parameters. To illustrate the practical applicability of ConditionNET, we implement an execution monitoring system on real hardware and validate its performance across different anomaly scenarios.

We consider several possibilities for future work. First, explaining the reasons behind the failure is important, as different actions could be taken by the robot depending on what happened. Second, we noticed that the model poorly generalizes to new robot embodiments not seen in the training dataset. To solve the domain gap between the training dataset and real-world robot implementation, and allow for better generalization to new robots, a large-scale manipulation dataset with robot failures would benefit the community.

\section*{Acknowledgments}
This work has been partially supported by the European Union project INVERSE under grant agreement No. 101136067.

\bibliographystyle{bibtex/IEEEtran}
\bibliography{main}

\end{document}